\documentclass{ecai}  % use option [doubleblind] for double blind submission and hiding the authors section

\usepackage{graphicx}
\usepackage{latexsym}
\usepackage{tabularx}
\usepackage{worldflags}
\usepackage{booktabs}
\usepackage[breaklinks=true]{hyperref}
\usepackage{xcolor}

\usepackage{tcolorbox}
\usepackage{pifont} % For dingbat symbols
\usepackage{tcolorbox}
\usepackage{pifont} % for \ding

\newenvironment{features}[1]
  {\begin{tcolorbox}[colback=green!10, colframe=green!60!gray, title=SOLVE-Med Demo Features]\par\vspace{0.1em}}
  {\end{tcolorbox}}

\begin{document}

\begin{frontmatter}

\title{SOLVE-Med: Specialized Orchestration for Leading Vertical Experts across Medical Specialties}

% ORCID
% \author[A]{\fnms{Roberta}~\snm{Di Marino}\orcid{0009-0005-7175-5325}\footnote{Equal contribution.}}
% \author[A]{\fnms{Giovanni}~\snm{Dioguardi}\orcid{0009-0006-5797-9485}\footnotemark}
% \author[A]{\fnms{Antonio}~\snm{Romano}\orcid{0009-0000-5377-5051}\footnotemark}
% \author[A]{\fnms{Giuseppe}~\snm{Riccio}\orcid{0009-0002-8613-1126}\footnotemark}
% \author[A]{\fnms{Mariano}~\snm{Barone}\orcid{0009-0004-0744-2386}\thanks{Corresponding Author. Email: mariano.barone@unina.it.}\footnotemark}
% \author[B]{\fnms{Marco}~\snm{Postiglione}\orcid{0000-0003-1470-8053}\footnotemark}
% \author[A]{\fnms{Vincenzo}~\snm{Moscato}\orcid{0000-0002-0754-7696}\footnotemark}

\author[A]{\fnms{Roberta}~\snm{Di Marino}\footnote{Equal contribution.}}
\author[A]{\fnms{Giovanni}~\snm{Dioguardi}\footnotemark}
\author[A]{\fnms{Antonio}~\snm{Romano}\footnotemark}
\author[A]{\fnms{Giuseppe}~\snm{Riccio}\footnotemark}
\author[A]{\fnms{Mariano}~\snm{Barone}\thanks{Corresponding Author. Email: mariano.barone@unina.it.}\footnotemark}
\author[B]{\fnms{Marco}~\snm{Postiglione}\footnotemark}
\author[A]{\fnms{Flora}~\snm{Amato}\footnotemark}
\author[A]{\fnms{Vincenzo}~\snm{Moscato}\footnotemark}

% use of \orcid{} is optional

\address[A]{University of Naples Federico II, DIETI, Naples, Italy}
\address[B]{Northwestern University, Department of Computer Science, Evanston, IL, United States}

\begin{abstract}
Medical question answering systems face deployment challenges including hallucinations, bias, computational demands, privacy concerns, and the need for specialized expertise across diverse domains. Here, we present SOLVE-Med, a multi-agent architecture combining domain-specialized small language models for complex medical queries. The system employs a Router Agent for dynamic specialist selection, ten specialized models (1B parameters each) fine-tuned on specific medical domains, and an Orchestrator Agent that synthesizes responses. Evaluated on Italian medical forum data across ten specialties, SOLVE-Med achieves superior performance with ROUGE-1 of 0.301 and BERTScore F1 of 0.697, outperforming standalone models up to 14B parameters while enabling local deployment. Our code is publicly available on GitHub: \url{https://github.com/PRAISELab-PicusLab/SOLVE-Med}.
\end{abstract}

\end{frontmatter}

\section{Introduction}

% QAS are widely adopted across various domains, yet providing accurate and context-aware answers remains particularly challenging in specialized fields such as medicine, where knowledge is complex, multidisciplinary, and often language-specific~\cite{ref_zhou}.
% %This issue is even more critical for the Italian medical context, where dedicated resources and language models are still limited~\cite{ref_medexpqa,ref_igea}. %

% While multi-agent approaches have been proposed to enhance specialization and scalability by distributing tasks among expert agents~\cite{ref_fouad,ref_guo}, existing QA solutions often rely on retrieval-based methods or prompt engineering, which struggle to achieve true modularity and specialization.

% In this demo, we present SOLVE-Med, a multi-agent framework for medical QA, designed to overcome these limitations. The system leverages a centralized orchestrator to coordinate several fine-tuned SLMs, each specialized in a specific medical field. Despite their reduced size, these agents contribute domain-specific expertise, enabling the orchestrator to produce accurate, coherent, and context-aware answers.

% The demo highlights the interactive capabilities of the system, showcasing how agent specialization and orchestration can support the generation of accurate medical responses. The approach aspires to reach a level of answer quality that could approach that of larger models, while offering advantages in modularity, scalability, and computational efficiency.

The rapid advancement of artificial intelligence in healthcare has catalyzed the emergence of sophisticated question-answering systems capable of addressing complex medical queries. With the Healthcare Chatbots Market projected to surpass a valuation of US\$ 11.8 billion by 2033\footnote{\href{https://www.globenewswire.com/news-release/2025/01/20/3011911/0/en/Healthcare-Chatbots-Market-is-Set-to-Surpass-Valuation-of-US-11-8-Billion-By-2033-Astute-Analytica.html}{https://www.globenewswire.com/news-release/2025/01/20/3011911/0/en/ Healthcare-Chatbots-Market}}, there is an urgent need for reliable, interpretable, and privacy-preserving medical AI systems that can operate effectively within clinical workflows. Large Language Models (LLMs) have demonstrated remarkable capabilities in medical question answering by enabling automatic access to complex knowledge across various domains \cite{singhal2025toward,zheng2025large}. However, their deployment in clinical contexts faces significant barriers, including hallucinations, bias, limited interpretability, high computational costs, and data privacy concerns associated with closed-source, cloud-based services \cite{nazi2024large,salam2025large}. Small Language Models (SLMs) represent a compelling alternative paradigm that addresses many of these challenges. By combining compact architectures with efficient fine-tuning on domain-specific data, SLMs require substantially fewer computational resources for both training and inference while enabling local deployment, reduced energy consumption, and enhanced privacy control \cite{chen2024role,wang2024comprehensive}. Importantly, despite their reduced scale, well-tuned SLMs can deliver high task performance and generate accurate, context-aware responses even in complex medical scenarios \cite{kim2025small}. This efficiency makes them particularly suitable for healthcare applications where resource constraints and data sensitivity are critical considerations. The multi-agent paradigm has gained considerable attention in question-answering systems, offering advantages in modularity, task separation, and system robustness \cite{bousetouane2025agentic,guo2024large}. In medical contexts, where queries often span multiple specialties and require diverse forms of expertise, this framework enables clearer functional boundaries and more reliable outputs \cite{yang2024llm,qiu2024llm}. However, existing approaches have not fully exploited the potential of combining specialized SLMs within a multi-agent architecture specifically designed for medical question answering. Here, we present SOLVE-Med (Specialized Orchestration for Leading Vertical Experts across Medical Specialties), a novel multi-agent system that addresses the specific demands of medical question answering through the strategic integration of domain-specialized SLMs. Our approach comprises three main components: (i) a Router Agent that performs multi-label classification to dynamically select relevant medical specialists based on query content, (ii) a pool of ten specialized SLMs, each fine-tuned on specific medical domains using Italian healthcare forum data gathered from Medicitalia\footnote{\url{https://www.medicitalia.it/}} and Dica33\footnote{\url{https://www.dica33.it/}}, and (iii) an Orchestrator Agent that synthesizes multiple expert responses into coherent, comprehensive answers. This architecture enables local deployment through compact models, significantly improving computational efficiency and preserving data privacy by eliminating dependence on external cloud infrastructure.

\begin{features}
    \noindent\ding{52}\quad The demo features a multi-turn chat interface, allowing users to submit queries and engage in an interactive dialogue with the system.
    
    \vspace{0.3em}\noindent\ding{52}\quad At each interaction step, SOLVE-Med presents not only the final response generated by the Orchestrator Agent but also the individual outputs from the selected medical specialists, along with the confidence scores assigned by the Router Agent.
    
    \vspace{0.3em}\noindent\ding{52}\quad Users are also allowed to direct queries to specific agents within the pool of medical specialists, enabling targeted exploration of domain-specific expertise.
\end{features}

\begin{figure*}[t]
    \centering
    \includegraphics[width=0.95\linewidth]{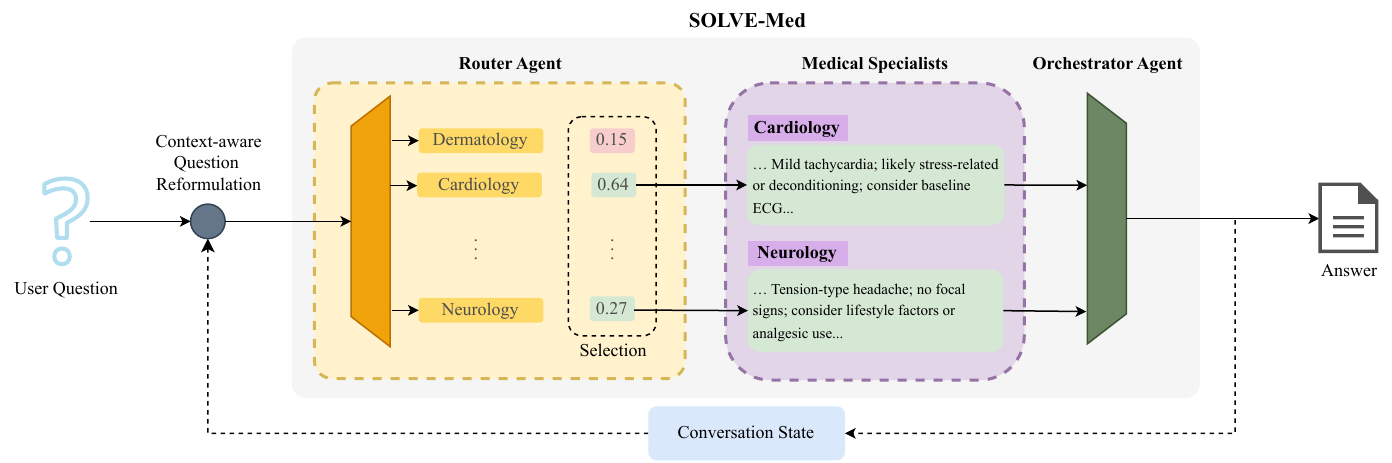}
    \caption{Overview of the SOLVE-Med architecture and multi-turn interaction flow. Given a user question, the system first performs context-aware reformulation based on the conversation state. The reformulated query is then processed by the Router Agent, which assigns relevance scores to domain-specific specialist models (SLMs). The most relevant specialists are selected to generate candidate responses, which are then integrated by the Orchestrator Agent into a single coherent medical answer. The conversation state is continuously updated to support multi-turn interactions.}
    \label{fig:architecture}
\end{figure*}

\section{SOLVE-Med: Overview}

% \begin{figure}[h]
%     \centering
%     \includegraphics[width=1\linewidth]{Architecture (1).pdf}
%     \caption{Architecture of SOLVE-Med. A user question is processed by the Router Agent, which selects the most relevant domain-specific specialists (SLMs). The selected specialists generate individual responses, which are then integrated by the Orchestrator Agent into a single, coherent medical answer.}
%     \label{fig:architecture}
% \end{figure}

% In this section, we present our proposed methodology for medical question answering \from{Marco}{G/R}{Da revisore, capisco che il problema che state risolvendo è quello del \textit{medical question answering}. è un po' semplicistico. Che aspetti del MQA risolve la vostra architettura? Si deve capire chiaramente ogni volta che parlate della vostra soluzione.}. As illustrated in Figure~\ref{fig:architecture}, the architecture is composed of three main components: (1) Router Agent (RA), (2) Medical Specialists, and (3) Orchestrator Agent (OA). The remainder of this section provides a detailed description of each component.
In this section, we present a detailed description of the SOLVE-Med architecture (see Figure~\ref{fig:architecture} for an overview), which comprises three core components. First, a \textit{Router Agent (RA)} dynamically selects the most suitable expert agents to address a given user query. Second, a pool of \textit{Medical Specialists}, each consisting of a small language model pre-trained in a specific domain of medical expertise, provides domain-specific responses. Third, an \textit{Orchestrator Agent (OA)} integrates the individual outputs from the specialists into a coherent and unified final response. 

% \paragraph{Router Agent.} The task of the RA is to select the most appropriate medical specialists based on the input query. Similar to real-world medical practice—\from{Marco}{G/R}{Le incidentali piacciono tanto agli LLM, ma non vengono usate così tanto nella pratica.}where a single case may require the expertise of multiple specialists—our architecture allows the RA to identify one or more doctors according to the nature of the question \from{Marco}{G/R}{Rileggete quanto scritto fino ad ora: sembra che il router agent si occupi di trovare effettivamente dei dottori}. The RA was implemented using a multi-label classifier, for which both traditional Machine Learning models (Logistic Regression, Random Forest, Support Vector Machine, XGBoost) and Transformer-based models (DistilBERT-base-multilingual-cased~\cite{ref_url1} were explored \from{Marco}{G/R}{Su quale dataset? Quali feature? }. For the final specialist selection, both threshold-based~\footnote{Only those specialists whose classification score exceeds a predefined threshold are selected\from{Marco}{G/R}{è un dettaglio importante: non va nelle footnote.}} and top-$n$~\footnote{The $n$ specialists with the highest classification scores are selected} approaches were tested. \from{Marco}{G/R}{L'ultima frase è molto poco chiara}

\paragraph{Router Agent} The Router Agent (RA) functions as a multi-label classifier that assigns one or more specialized medical agents to each input query. By routing questions to the most relevant specialists, the RA mimics the consultative nature of clinical workflows, enabling the system to provide multiple expert perspectives on complex queries. We implemented the RA by fine-tuning DistilBERT-base-multilingual-cased~\cite{distilbert-multilingual}, which supports rapid inference and low memory consumption, making it particularly suitable for deployment in resource-constrained medical environments. Training details are provided in Section~\ref{sec:expsetup}. Specialist selection is guided by the classifier's multi-label output, using one of these two strategies: 
\begin{itemize}
    \item \textit{Threshold-based}\; Medical specialists are selected based on whether their associated label scores exceed a fixed confidence threshold. In our experiments, thresholds were chosen to maximize the $F_\beta$ score, with $\beta = 2$ and $\beta = 3$, thus prioritizing recall over precision during this selection phase. This emphasis on recall reflects the system’s design: the Orchestrator Agent subsequently refines the response by filtering out incorrect or irrelevant contributions, thereby enhancing overall precision.
    \item \textit{Top-n-based}\; Medical specialists corresponding to the top-n highest-scoring labels are chosen. We evaluated SOLVE-Med's performance under top-2 and top-3 selection settings.
\end{itemize}

\paragraph{Medical Specialists} Once selected by the Router Agent, the medical specialists receive the user query and generate responses grounded in their respective areas of expertise. The specialist agents consist of a set of ten Small Language Models (SLMs), each fine-tuned on a distinct medical specialty using the LLaMA-3.2-1B-Instruct model~\cite{llama3.2-1b-instruct}. To further enhance runtime efficiency, we employed a quantized version of the model, significantly reducing memory consumption and inference latency while preserving acceptable response quality. Each SLM was trained to generate coherent, domain-specific answers with consistent style and terminology. Full training details are provided in Section~\ref{sec:expsetup}.

% \paragraph{Orchestrator Agent.} The OA is responsible for coordinating and synthesizing the responses provided by the selected specialists. It receives the user’s initial query and, after collecting the contributions from various experts, integrates them into a final response that is coherent, accurate, and comprehensive \from{Marco}{G/R}{è necessario presentare un esempio}. The OA was implemented using a LLM, specifically the Llama-3.1-8B-Instruct~\cite{ref_url2} model in its quantized version. The choice was driven by theoretical considerations—such as computational efficiency, semantic understanding, and coherence in handling multidisciplinary inputs\from{Marco}{G/R}{poco chiaro}—as well as experimental evidence supporting its performance \from{Marco}{G/R}{e dove sta questa experimental evidence?}. An alternative approach was also tested, where responses were weighted based on classifier confidence scores, but this did not yield improvements over an equal-weight fusion of contributions \from{Marco}{G/R}{Se dici che hai fatto una cosa, o me la mostri o penso che dici solo supercazzole e non solo non credo a quello che hai scritto, ma metto in discussione tutto il resto.}.

\paragraph{Orchestrator Agent} The Orchestrator Agent (OA) is responsible for aggregating and synthesizing the outputs of the selected medical specialists. It receives both the user’s original question and the set of responses generated by the relevant SLMs, and produces a single, integrated answer that is coherent, medically sound, and comprehensive. The OA is implemented using a quantized version of the Gemma-2-9B-IT model~\cite{gemma2-9b-it}. Its larger parameter count w.r.t. medical specialists enables effective integration of their contributions, helping to mitigate issues such as omissions or oversimplification that may arise when using smaller models. Despite its bigger size, quantization significantly reduces memory requirements and inference latency, ensuring the model remains suitable for deployment in hardware-constrained settings. The OA operates through a structured prompting strategy that frames it as a professional medical assistant. This prompt explicitly instructs the model to merge specialist outputs into a unified, evidence-based response that is clear, contextually appropriate, and aligned with the user’s original query.

\begin{table*}[t]
\centering
\caption{Comparison of model performances across various metrics.}
\begin{tabular}{lccccccccc}
    \toprule
    \textbf{Model/System} & \textbf{Rouge-1} & \textbf{Rouge-2} & \textbf{Rouge-L} & \textbf{Rouge-Lsum} & \textbf{BLEU} & \textbf{METEOR} & \textbf{BERT-P} & \textbf{BERT-R} & \textbf{BERT-F1} \\
    \midrule
    Llama-3.1-8B-Instruct         & 0.2183 & 0.0547 & 0.1484 & 0.1609 & 0.0210 & 0.2124 & 0.6395 & 0.6958 & 0.6657 \\
    Gemma-2-9b-it                & 0.2546 & 0.0649 & 0.1643 & 0.1925 & 0.0176 & 0.2422 & 0.6267 & 0.7070 & 0.6641 \\
    Velvet-14B                   & 0.2501 & 0.0662 & 0.1633 & 0.1801 & 0.0259 & 0.2527 & 0.6491 & 0.7094 & 0.6775 \\
    \midrule
    \textbf{SOLVE-Med (Top-2)}          & 0.2974 & 0.0760 & 0.1852 & 0.2159 & 0.0291 & \textbf{0.2660} & 0.6686 & 0.7250 & 0.6953 \\
    \textbf{SOLVE-Med (Top-3)}          & \textbf{0.3010} & 0.0748 & \textbf{0.1881} & \textbf{0.2197} & 0.0260 & 0.2655 & \textbf{0.6720} & \textbf{0.7256} & \textbf{0.6974} \\
    \textbf{SOLVE-Med (F\textsubscript{2})}    & 0.2920 & 0.0736 & 0.1828 & 0.2090 & 0.0269 & 0.2582 & 0.6645 & 0.7243 & 0.6928 \\
    \textbf{SOLVE-Med (F\textsubscript{3})}    & 0.2975 & \textbf{0.0765} & 0.1850 & 0.2144 & \textbf{0.0291} & 0.2654 & 0.6663 & 0.7247 & 0.6939 \\
    \bottomrule
\end{tabular}
\label{tab:combined_metrics_model_comparison}
\end{table*}

\section{Experiments}

In this section, we present a comparative evaluation of SOLVE-Med against other models and analyze the impact of the classifier on the system.

\subsection{Experimental Setup}\label{sec:expsetup}

\paragraph{Implementation Details} All fine-tuning and inference procedures involving the Medical Specialists and the Orchestrator Agent were conducted using Unsloth~\cite{unsloth2023}, which offers efficient resource utilization and accelerated training workflows. We adopted a Parameter-Efficient Fine-Tuning (PEFT) strategy~\cite{xu2023parameterefficientfinetuningmethodspretrained}, specifically employing the Low-Rank Adaptation (LoRA) technique~\cite{hu2021loralowrankadaptationlarge}. LoRA substantially reduces the number of trainable parameters while maintaining fine-tuning effectiveness.

\paragraph{Dataset} The training data for both the Router Agent (RA) and the SLMs was sourced from Italian medical forums. The original corpus consisted of approximately 700,000 question–answer pairs, each annotated with one of 102 platform-defined categories. To enhance semantic consistency and promote generalization, these fine-grained labels were manually consolidated into 10 macro-categories representing major medical domains:
\textit{Cardiology and Hematology, Dermatology and Aesthetics, Gastroenterology, Gynecology, General Medicine and Surgery, Neurology, Eye, ENT and Pulmonology, Orthopedics, Mental Health, and Urology and Andrology.}

To prevent data leakage between system components, we constructed disjoint training subsets for the Router Agent (RA) and the SLMs, ensuring strict separation of input data across modules. For each SLM, we selected 10,000 training, 500 validation, and 500 test samples by computing text embeddings for all question–answer pairs in the corpus. We then applied UMAP for dimensionality reduction~\cite{mcinnes2020umapuniformmanifoldapproximation}, followed by HDBSCAN clustering~\cite{McInnes2017}, to identify and sample the most representative QA pairs within each macro-category. For the RA, we sampled 5,000 training, 1,000 validation, and 1,000 test examples consisting of user queries paired with their corresponding macro-category labels. These examples were evenly distributed across the ten macro-categories to ensure balanced representation and support robust multi-label classification.

\paragraph{Evaluation Metrics} To evaluate the overall SOLVE-Med’s performance, we constructed a dedicated test set of 100 question–answer pairs, sampled from a portion of the source forum not used during model training. The dataset includes 10 manually selected questions per macro-category, chosen for clarity, medical relevance, and coverage of diverse clinical topics. The limited size reflects the manual curation effort and a focus on a high-quality evaluation.

System performance was assessed using a range of QA metrics. ROUGE \cite{ROUGE} and BLEU \cite{BLEU} were used to measure lexical overlap with reference answers, while METEOR \cite{METEOR} and BERTScore \cite{bertscore} captured semantic similarity, offering a more nuanced view of response quality. Router Agent's performance has been assessed in terms of its precision and recall.

\subsection{Results}
% Evaluations were also \from{Marco}{G/R}{"also" rispetto a cosa?} conducted on both the Router Agent classifier—measuring its accuracy in correctly routing questions to the appropriate medical specialties—and on the individual specialized SLMs, to assess the quality of their responses in isolation, before orchestration. This helped isolate the contribution of each component and confirm the effectiveness of the modular design.

\paragraph{Overall Performance} The evaluation of SOLVE-Med demonstrates promising performance across all measured dimensions. As shown in Table~\ref{tab:combined_metrics_model_comparison}, the system outperforms several competitive baselines, including larger standalone models such as LLaMA-3.1-8B-Instruct~\cite{llama3.1}, Gemma-2-9B-IT~\cite{gemma2-9b-it}, and Velvet-14B~\cite{velvet14b} (which is an Italian LLM). Moreover, a consistent trend emerges across classifier configurations: strategies that increase the average number of selected specialists tend to yield improved outcomes. This suggests that greater diversity among contributing domain experts enhances the completeness and informativeness of the final response, likely by capturing multiple relevant clinical perspectives.

\paragraph{Router Agent Evaluation} \label{sec:eva_RA}
To assess the influence of the Router Agent’s configuration on overall system behavior, we evaluated multiple label selection strategies using DistilBERT-base-multilingual-cased as the underlying multi-label classifier. For threshold-based selection, thresholds were tuned to maximize the F\textsubscript{$\beta$} score, using $\beta = 2$ and $\beta = 3$ to progressively prioritize recall over precision, yielding thresholds of \(0.15\) and \(0.10\), respectively. 

This reflects the design intent of the system, which favors broader coverage during specialist selection—delegating the responsibility of filtering irrelevant content to the Orchestrator Agent. As shown in Table~\ref{tab:router_eval}, increasing $\beta$ results in higher recall at the cost of reduced precision, as expected. Predicting a greater number of labels increases the likelihood of capturing the correct (ground-truth) category, but also introduces more false positives. This trade-off is particularly important given that the original dataset contains single-label annotations, whereas SOLVE-Med is designed to route queries to all potentially relevant specialists. The average number of specialists selected per query (\#Specialists) offers insight into this balance, quantifying the system’s emphasis on domain coverage versus specificity during the generation phase.

\begin{table}[t]
\centering
\caption{Router Agent performance under different label selection strategies. Higher recall comes at the cost of lower precision and more predicted labels per question.}
% \from{Marco}{G/R}{Le caption delle tabelle devono stare sopra o sotto? Verificare le guidelines}}
\begin{tabular}{lccc}
    \toprule
    \textbf{Strategy} & \textbf{Precision} & \textbf{Recall} & \textbf{\#Specialists} \\
    \midrule
    Top-2                & 0.4020 & 0.8040 & 2.000 \\
    Top-3                & 0.2960 & 0.8870 & 3.000 \\ \midrule
    F\textsubscript{2} threshold & \textbf{0.5699} & 0.8290 & 1.528 \\
    F\textsubscript{3} threshold & 0.4157 & \textbf{0.8900} & 2.277 \\
    \bottomrule
\end{tabular}
\label{tab:router_eval}
\end{table}

\section{Conclusion}

SOLVE-Med shows that a multi-agent architecture with specialized SLMs and orchestration outperforms a general-purpose model in response quality and control. The system confirms the feasibility of modular, efficient QA in low-resource settings. Future work includes human evaluations and improved context handling. The goal is to establish SOLVE-Med as a reliable support tool in clinical practice, without replacing medical judgment.

\begin{ack}
    This work was conducted with the financial support of (1) the PNRR MUR project PE0000013-FAIR and (2) the Italian ministry of economic development, via the ICARUS (Intelligent Contract Automation for Rethinking User Services) project (CUP: B69J23000270005).
\end{ack}

\bibliography{ecai}

\end{document}